\documentclass{llncs}

\usepackage{graphicx}
\usepackage{balance}
\usepackage{amsmath}	
\usepackage{algorithm}
\usepackage{algorithmic}
\usepackage{enumerate}

\newcommand{\figref}[1]{{Fig.\ref{#1}}}
\newcommand{\tabref}[1]{{Table~\ref{#1}}}

\def\vec#1{\mbox{\boldmath $#1$}}

\begin{document}
\frontmatter          % for the preliminaries
\pagestyle{headings}  % switches on printing of running heads
\mainmatter              % start of the contributions
\title{ An Analysis on Selection for High-Resolution Approximations in Many-objective Optimization}
\titlerunning{ An Analysis on Selection for High-Resolution Approximations in Many-objective Optimization}  
% abbreviated title (for running head)
%                                     also used for the TOC unless
%                                     \toctitle is used
%
\author{Hern{\'a}n Aguirre\inst{1} \and Arnaud Liefooghe\inst{2} \and S{\'e}bastien Verel\inst{3} \and Kiyoshi Tanaka\inst{1} }

\authorrunning{  Aguirre \and Liefooghe \and  Verel \and Tanaka}   
\institute{Faculty of Engineering, Shinshu University \\ 4-17-1 Wakasato, Nagano, 380-8553 JAPAN\\
\and
Universit{\'e} Lille 1 LIFL, UMR CNRS 8022, France\\
Inria Lille-Nord Europe, France\\ 
\and
Universit\'e du Littoral C\^ote d'Opale, LISIC, 62228 Calais, France\\
\email{ \{ahernan,ktanaka\}@shinshu-u.ac.jp} arnaud.liefooghe@lifl.fr ~ verel@univ-littoral.fr}

\maketitle              % typeset the title of the contribution

%---------------------------------------------------------------------------

\begin{abstract}

This work studies the behavior of three elitist multi- and many-objective evolutionary algorithms generating a high-resolution approximation of the Pareto optimal set. Several search-assessment indicators are defined to trace the dynamics of survival selection and measure the ability to simultaneously keep optimal solutions and discover new ones under different population sizes,  set as a fraction of the size of the Pareto optimal set. 

\end{abstract}

\section{Introduction}

In multi-objective optimization the aim of the optimizer is to find a good  approximation of the Pareto optimal set (POS) in terms of convergence and diversity of solutions. Convergence dictates that solutions in the approximation must be either members of the POS or close to it in objective space. Diversity usually implies that solutions in the approximation should be evenly spaced in objective space, following the distribution of the POS. 

In many-objective optimization, in addition to convergence and diversity, a third criterion also becomes a relevant aim of the optimizer. We call it the {\it resolution} of the approximation. The {\it resolution} is related to the number of points in the generated approximation of the POS. In many-objective problems the number of solutions in the POS increases exponentially \cite{Aguirre:04b} with the  dimensionality of the objective space. In general, many more points are required to cover uniformly with the same density a higher dimensional space. However, the required resolution of the approximation could vary depending on the application domain and the task of the optimization within the problem solving approach. A low resolution of the approximation may suffice in some domain applications. For example, domains where the formulation of the problem is already well understood and a solution has to be found and implemented regularly, such as the daily operational schedule of machines and the jobs assigned to them in a manufacturing plant. In these domains, the optimizer is often required to provide alternative exact solutions and too many of them could overwhelm a decision maker (operations manager) that must suggest a prompt course of action. In other application domains a high resolution of the approximation is required. For example in design optimization, where the problem-solving cycle often starts with multiple, sometimes ill defined, problem formulations and uses  optimization as a tool to validate the understanding of the problem and to discover new features about it. In these application domains it is not unusual to require that the optimizer provides approximations of the POS with tens of thousands or even hundreds of thousands of solutions. These approximations are subjected to data mining and analysis to verify and improve the problem formulation itself, understand the tradeoffs between variables and objectives, and extract valuable design knowledge \cite{Nishio:2014}. Thus, a many-objective optimizer should also aim to find an approximation with a {\it resolution} that properly captures the POS, with enough points to provide a useful description of it, depending on the dimensionality of the objective space and the optimization task at hand.

Many-objective optimization was initially attempted using evolutionary algorithms that proved effective for two and three objectives only to discover their lack of scalability.  A significant part of the research effort has been understanding the reasons for their failure and improving them, particularly in terms of convergence. Recently, some many-objective optimizers are being proposed \cite{Hadka:2013,Aguirre:2013b}. However, the performance of the improved and newly proposed algorithms is commonly assessed using a relatively very small number of points focusing mostly on convergence and/or diversity.  The resolution of the approximation in many-objective optimization has not been deeply studied and it is not clear the capabilities and behavior of the algorithms under this additional important criterion.   

In this work we analyze the behavior of three elitist multi- and many-objective evolutionary algorithms generating a high-resolution approximation of the POS. We define a basic indicator for resolution, the accumulated gain of the population, and several generational search-assessment indices respect to the POS. We trace the dynamics of survival selection and study the ability to simultaneously keep Pareto optimal (PO) solutions in the population and find new ones to improve the resolution of the approximation, setting population size as a fraction of the size of the POS. We use MNK-landscapes with $3-6$ objectives and $20$ bits, for which it is possible to know by enumeration all PO solutions.

\section{Methodology} \label{sec:proposed-method}

An important objective of this work is to analyze the ability of the algorithms to generate a high-resolution approximation of the POS.  A simple and basic indicator for resolution is to count the number of PO solutions found by the algorithms. In many-objective problems is likely that the population size is considerable smaller than the size of the POS. Thus, to achieve a good resolution  the algorithms should first be able to hit the POS with some members of the population and then continue discovering other PO solutions. The ability to converge towards the POS is a very important feature of the algorithm. In this study, we focus our attention mostly on the ability of the algorithm to continue discovering new PO solutions assuming that the algorithms can converge to the POS.

To evaluate this ability we use four MNK-landscapes \cite{Aguirre:04b} randomly generated with $M=3$, $4$, $5$, $6$ objectives, $N=20$ bits, and $K=1$ epistatic bit. In small landscapes with low non-linearity it is relatively simple for the algorithm to hit the optimal set. It is also possible to enumerate them and know the POS in order to analyze the dynamic of the algorithms respect to the optimum set. The exact number of PO solutions found by enumeration and the number of non-dominated fronts are shown in \tabref{POST} under columns $|POS|$ and $Fronts$, respectively. The same table also shows the corresponding fraction ($\%$) of the population sizes $|P|$ to the $|POS|$  for various population sizes investigated here.

%%%%%%%%%%%%%%%%%%%%%%%%%%%%%%%%%%% 

\begin{table}[tdp]
\caption{Number of Pareto optimal solutions $|POS|$ and number of non-dominated {\it Fronts} in the landscapes with $M=3$, $4$, $5$, and $6$ objectives. Also, fraction of $|P|$ / $|POS|$  (in $\%$) for various population sizes $|P|$ investigated in this study. }
\begin{center}
\begin{tabular}{c|c|c|rrrrrrrrr}
\hline \hline
& & & \multicolumn{9} { c }{ $|P|$ / $|POS|$ (\%)}\\
$M$ & $|POS|$ & $Fronts$ & ~~~50 & ~~100 & ~~200 & ~~500 &~1,000 & ~2,000 & ~4,000 & ~5,600 & 11,200\\
\hline
3 & 152 & 258 & 32.9 & 65.8 & 132.6 & & & & & & \\
4 & 1,554 & 76 & 3.2 & 6.4 & 12.9 & 32.2 & 64.4 & & & \\
5 & 6,265 & 29 & 0.8 & 1.6 & 3.2 & & & 31.9 & 63.8 &	& \\
6 & 16,845 & 22 & 0.3& 0.6 & 1.2 & & &	 & & 33.2 &66.5\\ 
\hline
\end{tabular}
\end{center}
\label{POST}
\end{table}%

%%%%%%%%%%%%%%%%%%%%%%%%%%%%%%%%%%% 

We run the algorithms for a fixed number of $T$ generations, collecting in separate files the sets of non-dominated solutions $\mathcal{F}_1(t)$ found at each generation. The approximation of the POS for a run of the algorithm, denoted $\mathcal{A}(T)$, is built by computing the non-dominated set from all generational non-dominated sets $\mathcal{F}_1(t)$, $t=0,1,\cdots,T$, making sure no duplicate solutions are included.  In general, the approximation at generation $t$ is given by
\begin{equation}
\mathcal{A}(t) = \{ \vec{x}  :   \vec{x}  \in \mathcal{X}(t) =  \mathcal{A}(t-1) \cup \mathcal{F}_1(t) \setminus  \mathcal{A}(t-1) \cap  \mathcal{F}_1(t)  \land \not\exists \vec{y} \in \mathcal{X}(t) ~ \vec{y} \succeq \vec{x} \} 
\end{equation}
\begin{equation}
\mathcal{A}(0) =  \mathcal{F}_1(0),
\end{equation}
where $\vec{y} \succeq \vec{x}$ denotes solution $\vec{y}$ Pareto dominates solution $\vec{x}$. The basic resolution index $\alpha$ of the approximation at generation $t$ is,
\begin{equation}
\alpha(t) = \frac{|\{ \vec{x} : \vec{x} \in \mathcal{A}(t) \land \vec{x} \in POS \} |}{|POS|},
\end{equation}
which gives the fraction of the accumulated number of PO solutions found until generation $t$ to the size of the POS. The highest resolution of the generated approximation of the POS is achieved when all Pareto optimal solutions are found.  Similarly, the accumulated population gain at generation $t$ can be expressed as 
\begin{equation}
\beta(t) = \frac{|\{ \vec{x} : \vec{x} \in \mathcal{A}(t) \land \vec{x} \in POS \} |}{|P|}.
\end{equation}

For our analysis on the dynamics of the algorithm, we compare the sets $\mathcal{F}_1(t)$ with the POS to determine which solutions in $\mathcal{F}_1(t)$ are Pareto optimal and compute several generational search-assessment indices ${I_t}$  (${\tau}_t$, ${\tau}^*_t$, ${\tau}^+_t$, ${\tau}^-_t$, ${\delta}_t$, ${\gamma}_t$), as shown in \tabref{INDEXES}. Note that these generational indexes are expressed as a fraction of the population size $|P|$. In this work we analyze them and their average value $\bar{I}$ ($\bar {\tau}$, $\bar {\tau}^*$, $\bar {\tau}^+$, $\bar {\tau}^-$, $\bar {\delta}$, $\bar {\gamma}$) taken over all generations computed as $\bar{I} = \frac{1}{T+1}\sum_{t=0}^{T} I_t $.

%%%%%%%%%%%%%%%%%%%%%%%%%%%%%%%%%%% 
\begin{table}[tdp]
\caption{Generational search-assessment indices $I_t$. Measures are taken on non-dominated population $\mathcal{F}_1(t)$ with respect to $\mathcal{F}_1(t-1)$ and/or the POS, normalized by population size $|P|$.}
\begin{center}
\begin{tabular}{c|l|l}
\hline \hline
$I_t$ & \hspace{3cm} Formula  & Comment \\
\hline 
$\tau_t$ &  $|\{ \vec{x}  :   \vec{x}  \in \mathcal{F}_1(t) \land \vec{x} \in POS\}| ~/~ |P| $ \quad & PO solutions  \\
$\tau^-_t$ & $|\{ \vec{x}  :   \vec{x}  \in \mathcal{F}_1(t) \land \vec{x} \in \mathcal{F}_1(t-1) \land \vec{x} \in POS\}| ~/~ |P|$ \quad & Old PO solutions   \\
$\tau^+_t$  &  $|\{ \vec{x}  :   \vec{x}  \in \mathcal{F}_1(t) \land \vec{x} \not\in \mathcal{F}_1(t-1) \land  \vec{x} \in POS \}| ~/~|P| $ & Possibly new PO solutions  \\
$\tau^*_t$ &  $| \{ \vec{x} :  \vec{x}  \in \mathcal{F}_1(t) \land \vec{x} \not\in \cup_{k=1}^{t-1}\mathcal{F}_1(k) \land \vec{x} \in POS \} | ~/ ~|P| $ & Absolutely new PO solutions \\
$\delta_t$  & $|\{ \vec{x}  :   \vec{x}  \in \mathcal{F}_1(t-1)  \land \vec{x} \not\in \mathcal{F}_1(t) \land \vec{x} \in POS \}| ~/~|P| $ & Dropped PO solutions \\
$\gamma_t$    & $|\{ \vec{x} :  \vec{x}  \in \mathcal{F}_1(t)   \land \vec{x} \not\in POS\} | ~/~|P|$ & Non-dominated, not PO sol.\\
\end{tabular}
\end{center}
\label{INDEXES}
\vspace{-0.5cm}
\end{table}%
%%%%%%%%%%%%%%%%%%%%%%%%%%%%%%%%%%% 

In this work we analyze NSGA-II \cite{bib:Deb00}, IBEA \cite{Zitzler:04}, and the Adaptive { $\varepsilon$-Sampling and  $\varepsilon$-Hood algorithm} \cite{Aguirre:2013b}.  In the following we briefly describe these algorithms, particularly fitness assignment, survival selection, and parent selection.

\section{Algorithms}

\subsection{NSGA-II}

NSGA-II is an elitist algorithm that uses Pareto dominance and crowding  estimation of solutions for survival and parent selections. To compute fitness of the individuals, the algorithm joins the current population $\mathcal{P}_t$ with its offspring $\mathcal{Q}_t$ and divide it in non-dominated fronts $\mathcal{F}=\{\mathcal{F}_i\},i=1,2,\cdots,N_F$ using the non-dominated sorting procedure. It also calculate the crowding distance $d_j$ of solutions within the fronts $\mathcal{F}_i$. The fitness of $j$-th solution in the $i$-th front is $\text{Fitness} (\vec{x}_j) = (i, d_j)$, where the front number is the primary rank and crowding distance the secondary rank.  Survival selection is performed by copying iteratively the sets of solutions $\mathcal{F}_i$ to the new population $\mathcal{P}_{t+1}$  until it is filled. If the set $\mathcal{F}_i$, $i > 1$, overfills the new population $\mathcal{P}_{t+1}$, the required number of solutions are chosen based on their secondary rank $d_j$. Parent selection for reproduction consists of binary tournaments between randomly chosen individuals from  $\mathcal{P}_{t+1}$ using their primary rank $i$ to decide the winners, breaking ties with their secondary rank $d_j$.

\subsection{IBEA (Indicator-Based Evolutionary Algorithm)}
%==========================
The main idea of IBEA \cite{Zitzler:04} is to introduce a total order between solutions by means of an arbitrary binary quality indicator $I$. The fitness assignment scheme of IBEA is based on a pairwise comparison of solutions in a population with respect to indicator~$I$. Each individual $\vec{x}$ is assigned a fitness value measuring the ``loss in quality'' if $\vec{x}$ was removed from the population $P$, i.e., $\text{Fitness}(\vec{x}) = \sum_{\vec{x}' \in P\setminus \{\vec{x}\}} ( -e^{-I(\vec{x}',\vec{x})/\kappa} )$, where $\kappa > 0$ is a user-defined scaling factor. Survival selection is based on an elitist strategy that combines the current population $\mathcal{P}_t$ with its offspring $\mathcal{Q}_t$, iteratively deletes worst solutions until the required population size is reached, and assigns the resulting population to $\mathcal{P}_{(t+1)}$. Here, each time a solution is deleted the fitness values of the remaining individuals are updated.  Parent selection for reproduction consists of binary tournaments between randomly chosen individuals using their fitness to decide the winners.

Different indicators can be used within IBEA.
We here choose to use the binary additive $\epsilon$-indicator~($I_{\epsilon+}$), as defined by the original authors \cite{Zitzler:04}.
\begin{equation}
\label{eq:eps}
I_{\epsilon +} (\vec{x}, \vec{x}') = \max_{i \in \{1,\dots,n\}} \{ f_i(\vec{x}) - f_i(\vec{x}') \}
\end{equation}
$I_{\epsilon+}(\vec{x},\vec{x}')$ gives the minimum value by which a solution~$\vec{x} \in P$ has to, or can be translated in the objective space in order to weakly~dominate another solution~$\vec{x}' \in P$. More information about IBEA can be found in \cite{Zitzler:04}.

\subsection{The {A$\varepsilon$S$\varepsilon$H} }

Adaptive  $\varepsilon$-Sampling and  $\varepsilon$-Hood (A$\varepsilon$S$\varepsilon$H) \cite{Aguirre:2013b} is an elitist evolutionary many-objective algorithm that applies $\varepsilon$-dominance principles for survival selection and parent selection.  There is not an explicit fitness assignment method in this algorithm. 

Survival selection joins the current population $\mathcal{P}_t$ and its offspring $\mathcal{Q}_t$ and divide it in non-dominated fronts $\mathcal{F}=\{\mathcal{F}_i\},i=1,2,\cdots,N_F$ using the non-dominated sorting procedure. 
In the rare case where the number of non-dominated solutions is smaller than the population size $|\mathcal{F}_1| < |P|$, the sets of solutions $\mathcal{F}_i$ are copied iteratively to $\mathcal{P}_{t+1}$ until it is filled; if set $\mathcal{F}_i$, $i > 1$, overfills $\mathcal{P}_{t+1}$, the required number of solutions are chosen randomly from it.  On the other hand, in the common case where $|\mathcal{F}_1| > |P|$, it calls {\it $\varepsilon$-sampling} with parameter $\varepsilon_s$. This procedure iteratively samples randomly a solution from the set  $\mathcal{F}_1$, inserting the sample in $\mathcal{P}_{t+1}$ and eliminating from $\mathcal{F}_1$ solutions $\varepsilon$-dominated by the sample. After sampling, if $\mathcal{P}_{t+1}$ is overfilled solutions are randomly eliminated from it. Otherwise,  if there is still room in $\mathcal{P}_{t+1}$, the required number of solutions are randomly chosen from the initially $\varepsilon$-dominated solutions and added to $\mathcal{P}_{t+1}$. 

After survival selection there is not an explicit ranking that could be used to bias mating. Rather, for parent selection the algorithm first uses a procedure called {\it $\varepsilon$-hood creation} to cluster solutions in objective space. This procedure randomly selects an individual from the surviving population and applies $\varepsilon$-dominance with parameter $\varepsilon_h$. A neighborhood is formed by the selected solution and its $\varepsilon_h$-dominated solutions. Neighborhood creation is repeated until all solutions in the surviving population have been assigned to a neighborhood. Parent selection is implemented by the procedure {\it $\varepsilon$-hood mating}, which sees neighborhoods as elements of a list than can be visited one at the time in a round-robin schedule. The first two parents are selected randomly from the first neighborhood in the list.  The next two parents will be selected randomly from the second neighborhood in the list, and so on. When the end of the list is reached, parent selection continues with the first neighborhood in the list. Thus, all individuals have the same probability of being selected within a specified neighborhood, but due to the round-robin schedule individuals belonging to neighborhoods with fewer members have more reproduction opportunities that those belonging to neighborhoods with more members.

Both epsilon parameters $\varepsilon_s$ and $\varepsilon_h$ used in survival selection and parent selection, respectively, are dynamically adapted during the run of the algorithm.  Further details about A$\varepsilon$S$\varepsilon$H can be found in \cite{Aguirre:2013b}.

%%%%%%%%%%%%%%%%%%%%%%%%%%%%%%%%%%% 

\section{Experimental Results and Discussion}
\subsection{ Operators of Variation and Parameters }

In all algorithms we use two point crossover with rate $pc=1.0$, and bit flip mutation with rate $pm=1/n$. In A$\varepsilon$S$\varepsilon$H we set the reference neighborhood size $H^{Ref}_{size}$ to 20 individuals. The mapping function $\vec{f}(\vec{x}) \mapsto^{\epsilon} \vec{f} ^{'}(\vec{x})$ used for $\varepsilon$-dominance in $\varepsilon$-sampling truncation and $\varepsilon$-hood creation is additive, $f^{'}_i = f_i + \varepsilon,  i=1, 2, \cdots, m$. For IBEA, the scaling factor is set to $\kappa=0.001$. The algorithms run for $T=100$ generations.  Results analyzed here were obtained from 30 independent runs of the algorithms.

\subsection{Accumulated number of Pareto optimal solutions found}

\begin{figure}[t]
\begin{minipage}{0.49\textwidth}
  \centering
  \includegraphics[width=0.8\linewidth,keepaspectratio=true, angle=270]{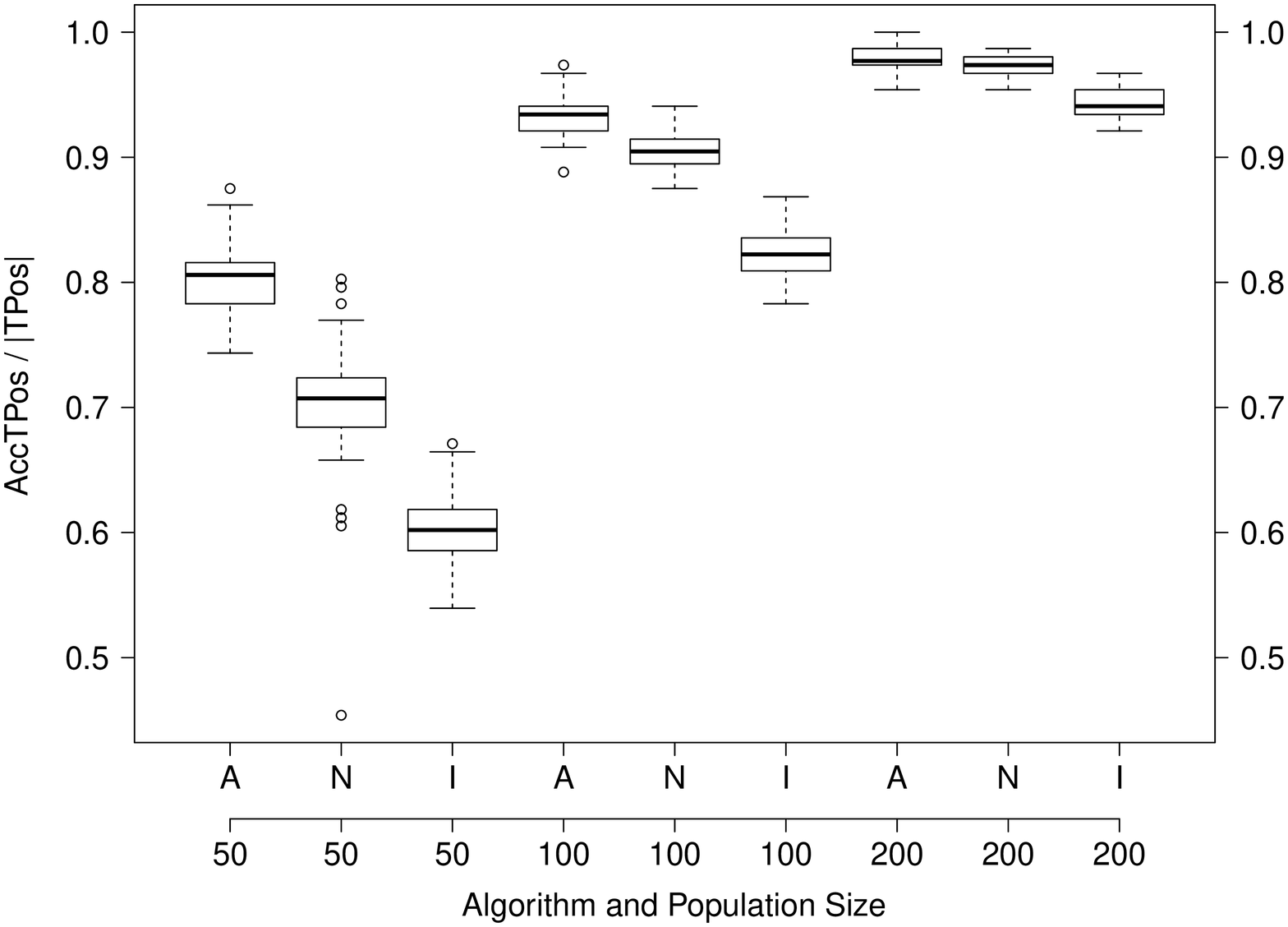}
\begin{center} (a) $M$=3 objectives \end{center}
\end{minipage} \hspace{2mm}
\begin{minipage}{0.48\textwidth}
  \centering
  \includegraphics[width=0.8\linewidth,keepaspectratio=true, angle=270]{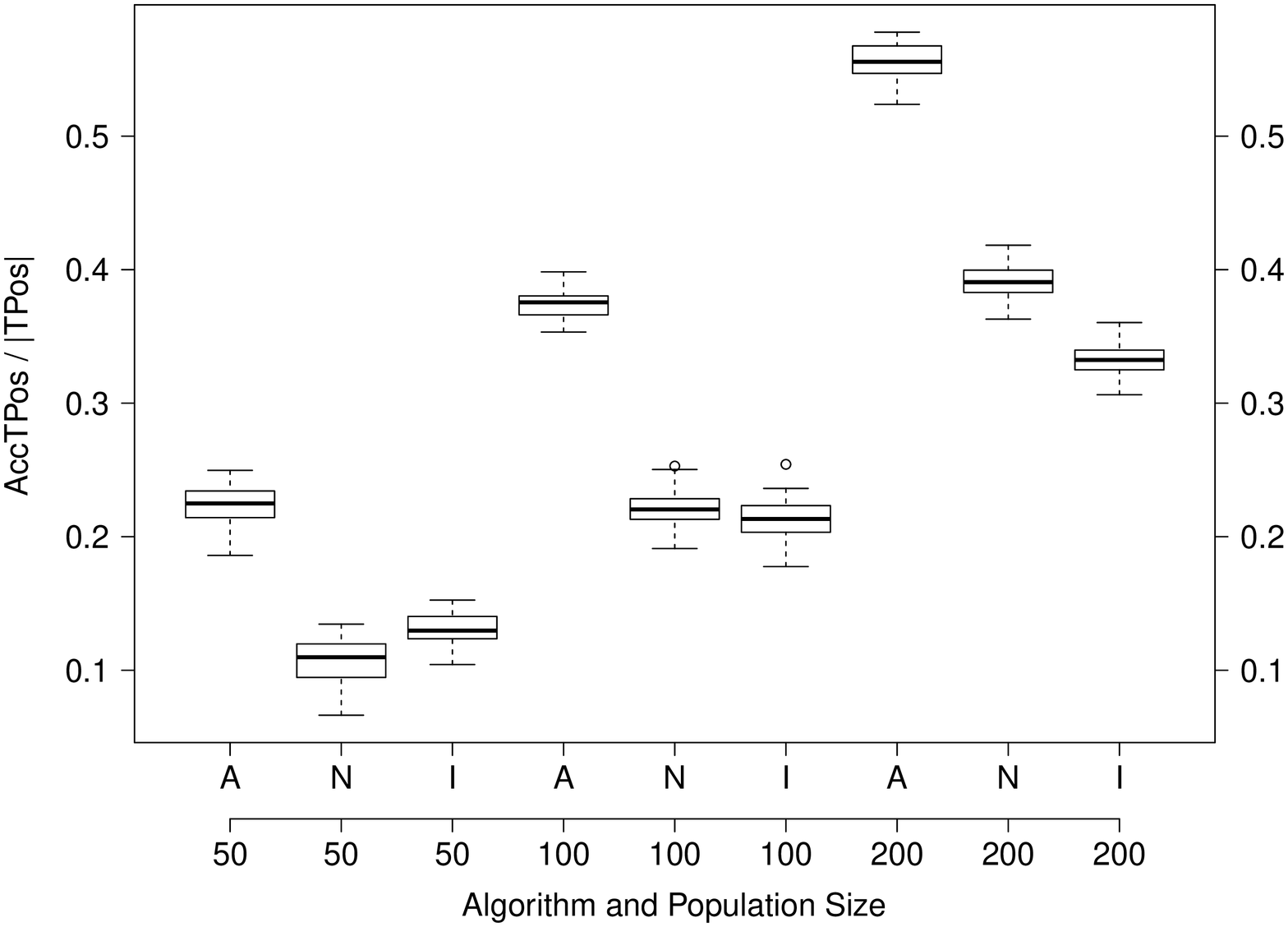}
\begin{center} (b) $M$=4 objectives \end{center}
\end{minipage} 
\begin{minipage}{0.48\textwidth}
  \centering
  \includegraphics[width=0.8\linewidth,keepaspectratio=true, angle=270]{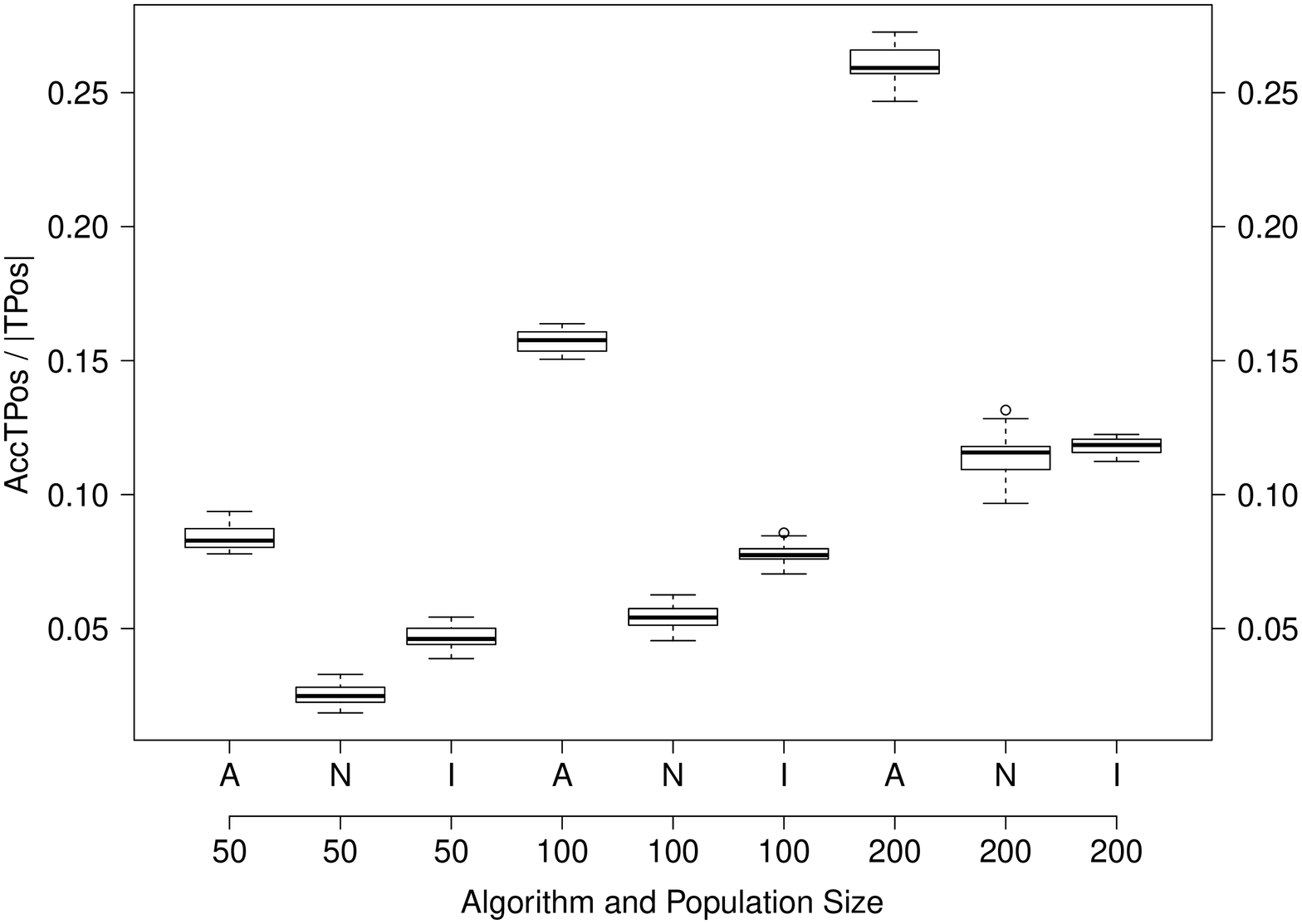}
\begin{center} (c) $M$=5 objectives \end{center}
\end{minipage} \hspace{2mm}
\begin{minipage}{0.48\textwidth}
  \centering
  \includegraphics[width=0.8\linewidth,keepaspectratio=true, angle=270]{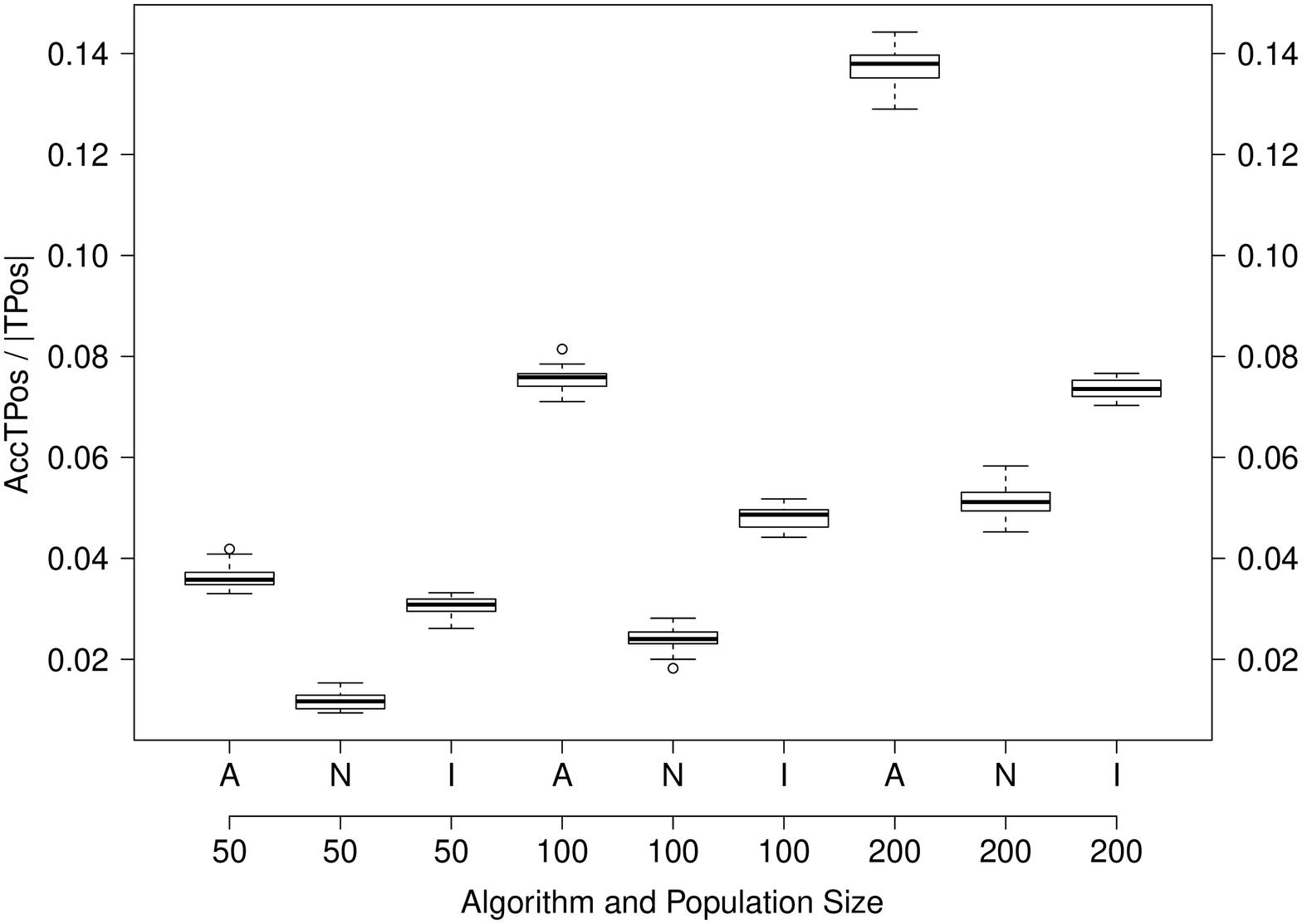}
\begin{center} (d) $M$=6 objectives \end{center}
\end{minipage}
  \caption{ Resolution of the approximation at the end of the run $\alpha(T)$, i.e. ratio of accumulated number of Pareto optimal solutions found to the size of the POS. Population sizes 50, 100, and 200 for 3, 4, 5, and 6 objectives. Algorithms A$\varepsilon$S$\varepsilon$H (A), NSGA-II (N) and IBEA (I).}
  \label{fig:R_AFTPos_TPos_P50100200}
  \vspace{-0.0cm}
\end{figure}

\begin{figure}[t]
\begin{minipage}{0.48\textwidth}
  \centering
  \includegraphics[width=0.8\linewidth,keepaspectratio=true, angle=270]{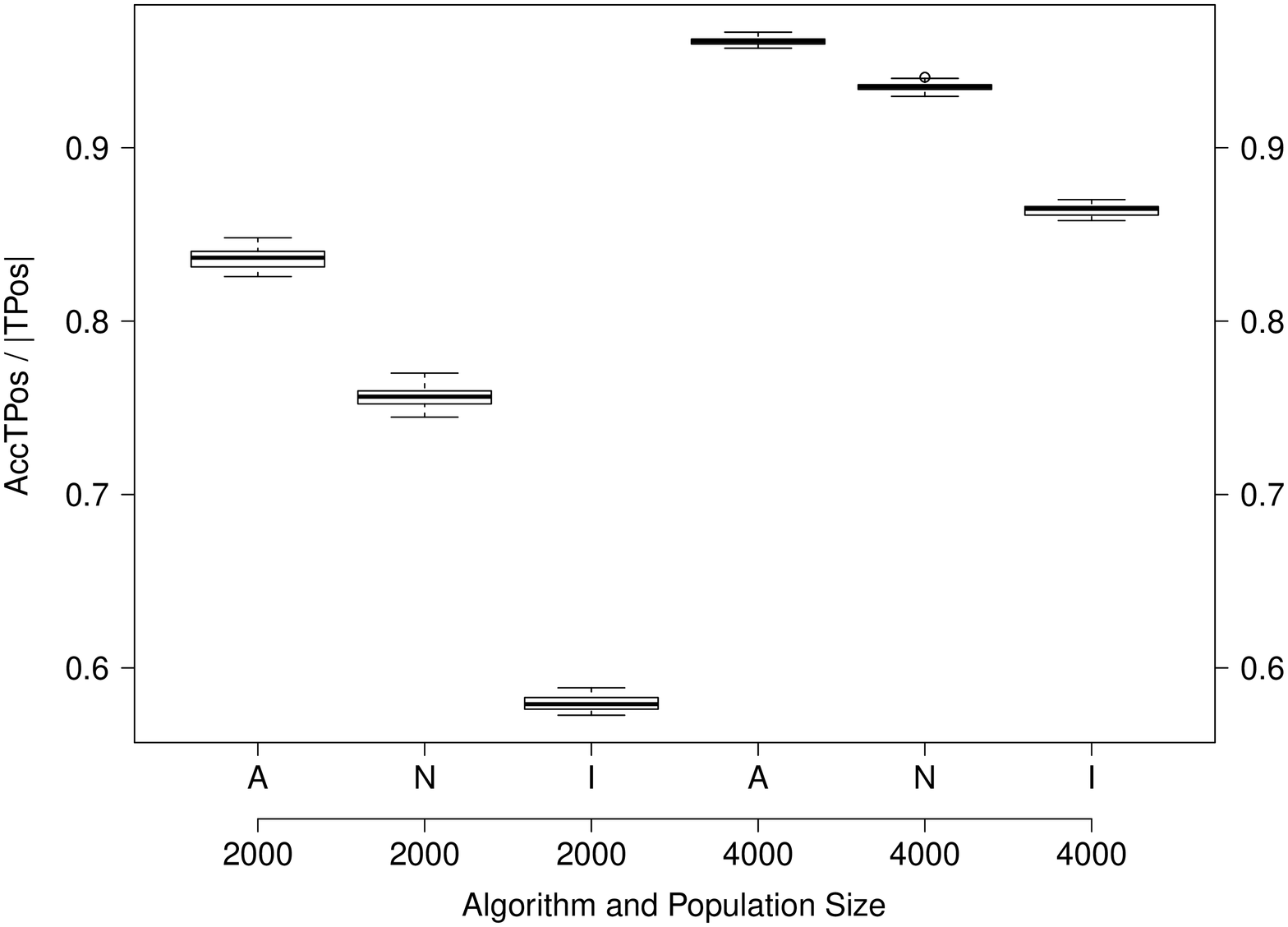}
\begin{center} (a) $M$=5 objectives \end{center}
\end{minipage} 
\begin{minipage}{0.48\textwidth}
 \vspace{1mm}
  \centering
  \includegraphics[width=0.8\linewidth,keepaspectratio=true, angle=270]{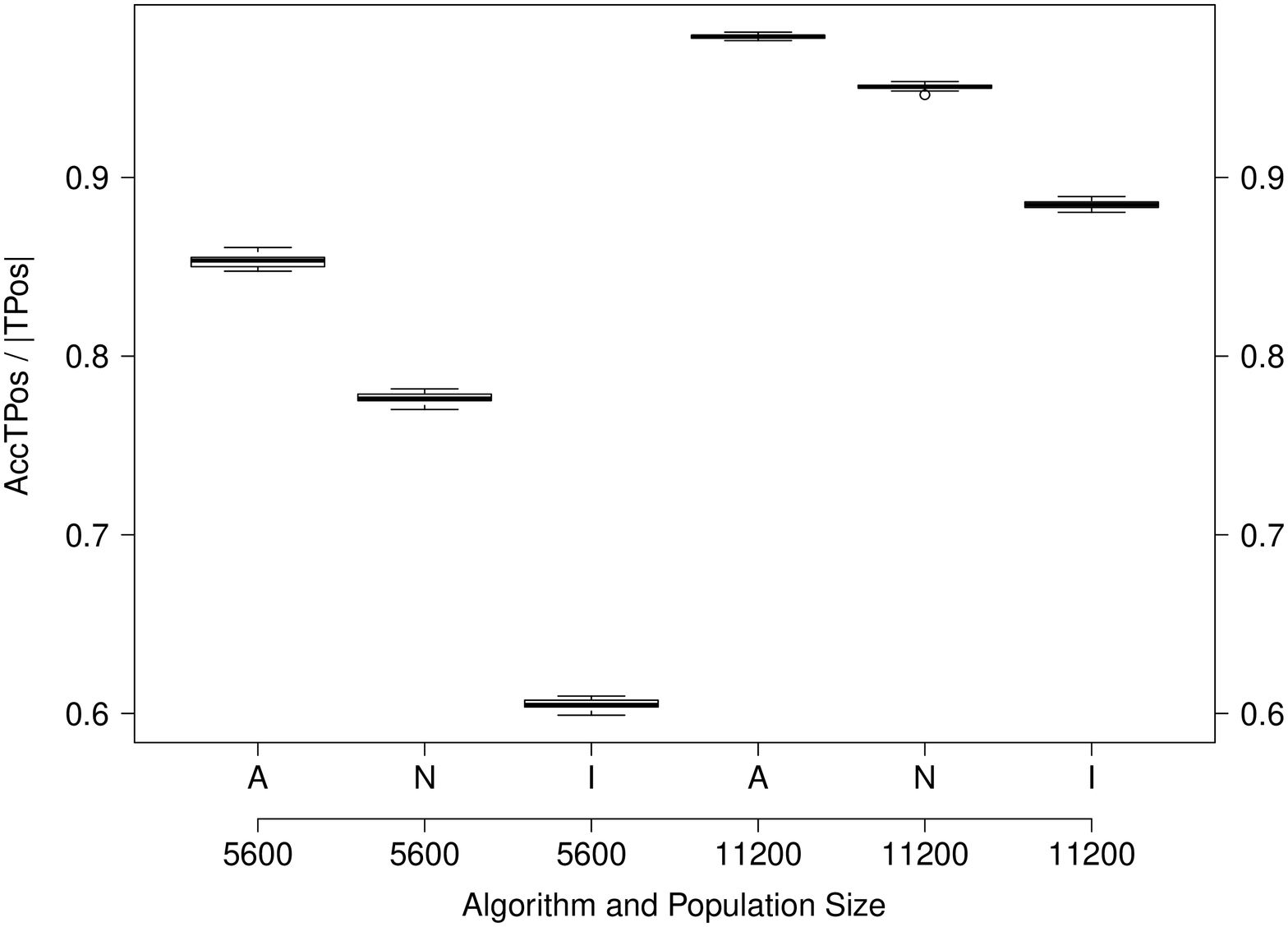}
 % \vspace{2mm}
\begin{center} (b) $M$=6 objectives \end{center}
\end{minipage}
  \caption{ Ratio of accumulated number of Pareto optimal solutions found to the size of the POS. Population sizes $\{500,1000\}$, $\{2000,4000\}$, and $\{5600, 11200\}$ for 4, 5, and 6 objectives, respectively. Algorithms A$\varepsilon$S$\varepsilon$H (A), NSGA-II (N) and IBEA (I). }
  \label{fig:R_AFTPos_TPos_P1323}
  \vspace{-0.0cm}
\end{figure}

\figref{fig:R_AFTPos_TPos_P50100200} shows the the basic resolution index $\alpha(T)$ of the approximation at the end of the run, i.e. the  ratio of accumulated number of PO solutions found to the size of the POS. Results are shown for $3$, $4$, $5$, and $6$ objectives using population sizes of $\{50,100,200\}$. Similarly, \figref{fig:R_AFTPos_TPos_P1323} shows results for  $5$, and $6$ objectives using larger populations sizes, between $500$ and $11,2000$ individuals. 

Note that  A$\varepsilon$S$\varepsilon$H finds many more Pareto optimal solutions than NSGA-II and IBEA for all population sizes and number of objectives tried here, whereas NSGA-II finds more solutions than IBEA when population sizes are relatively a large fraction of the size of the POS. See \figref{fig:R_AFTPos_TPos_P50100200} (a) and \figref{fig:R_AFTPos_TPos_P1323} (a)-(b) where population sizes correspond roughly to $33\%$, $66\%$, and $133\%$ of the POS for 3 objectives and    $33\%$ and $66\%$ for 5 and 6 objectives, as shown in $\tabref{POST}$.  On the contrary, IBEA finds more solutions than NSGA-II when population sizes are relatively a small fraction of the POS.  See \figref{fig:R_AFTPos_TPos_P50100200} (c)-(d) where population sizes $\{50,100, 200\}$ are used in $5$ and $6$ objectives, which correspond to fractions in the range $0.3\%-3.2\%$ of the POS. In $4$ objectives, \figref{fig:R_AFTPos_TPos_P50100200} (b), an interesting transition can be observed. When the smallest population is used, i.e. 50 individuals $\sim 3.2\%$ of POS, IBEA finds more solutions than NSGA-II. For a population size of 100 $\sim 6.4\%$ of POS NSGA-II finds a slightly larger number of solutions than IBEA.  For a population size of 200 $\sim 12.9\%$ of POS, NSGA-II finds a significant larger number of solutions than IBEA. 

The gap between A$\varepsilon$S$\varepsilon$H and the other two algorithms augments when the population size increases within a range in which it still is a small fraction of the POS, as shown in \figref{fig:R_AFTPos_TPos_P50100200} (b)-(d) where the ranges in which population size increase are $3.2\%-12.9\%$, $0.8\%-3.2\%$, and $0.3\%-1.2\%$ of POS for $4$, $5$, and $6$ objectives, respectively.  On the other hand, the gap reduces when population size increases within a range in which it is a large fraction of the POS, as shown in  \figref{fig:R_AFTPos_TPos_P50100200} (a) and  \figref{fig:R_AFTPos_TPos_P1323} (a)-(b) where the ranges in which population size increase are roughly  $33\%-133\%$ of  POS for $3$ objectives and $33\%-66\%$ for $5$ and $6$ objectives.  
%%%%%%%%%%%%%%%%%%%%%%%%%%%%%%%%%%% 

\begin{figure}[t]
\begin{minipage}{0.48\textwidth}
  \centering
  \includegraphics[width=0.8\linewidth,keepaspectratio=true, angle=270]{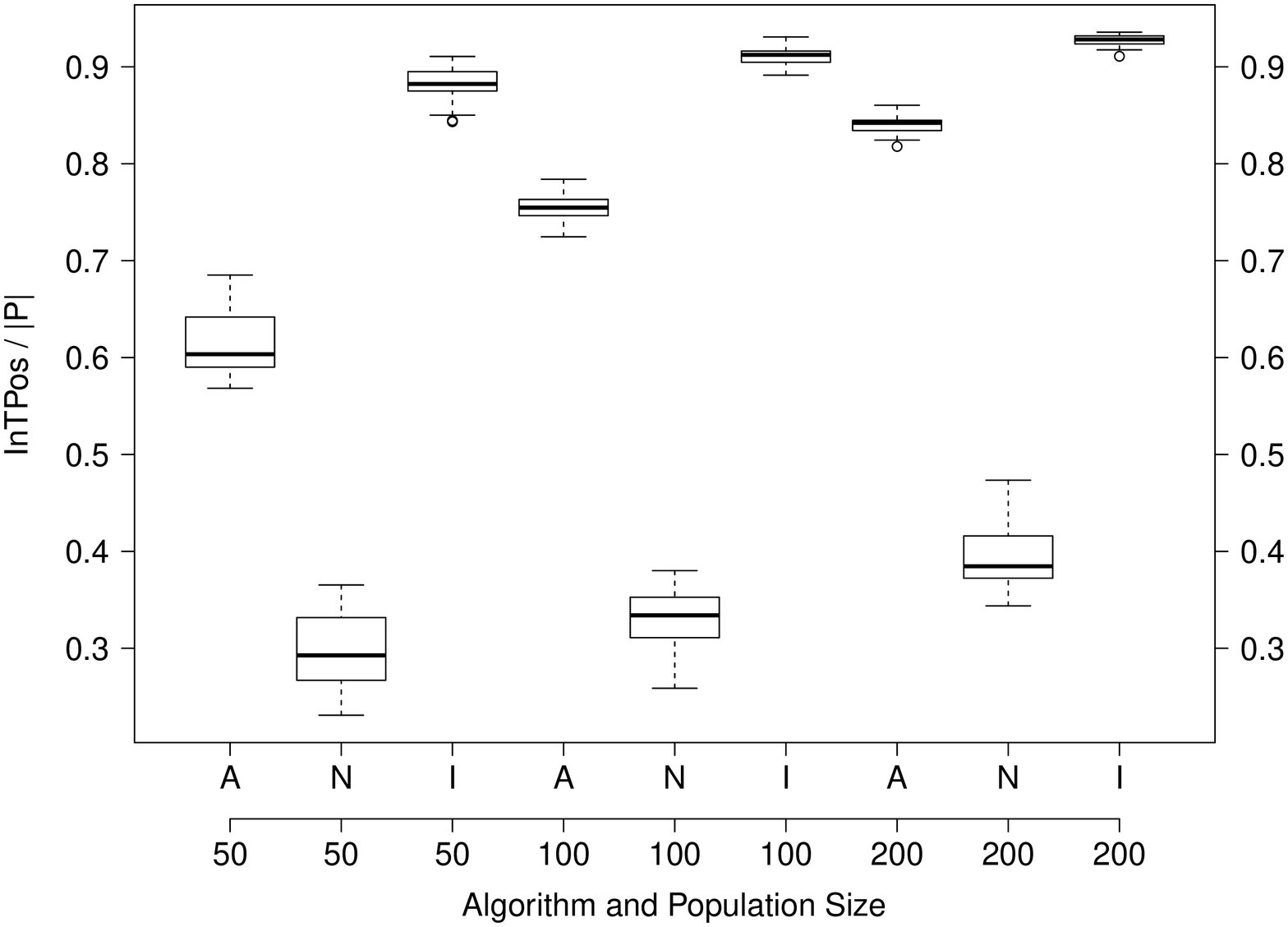}
  \begin{center} (a)  $\bar {\tau}$: Pareto optimal \end{center}
\end{minipage}
\hspace{2mm}
\begin{minipage}{0.48\textwidth}
  \centering
  \includegraphics[width=0.8\linewidth,keepaspectratio=true, angle=270]{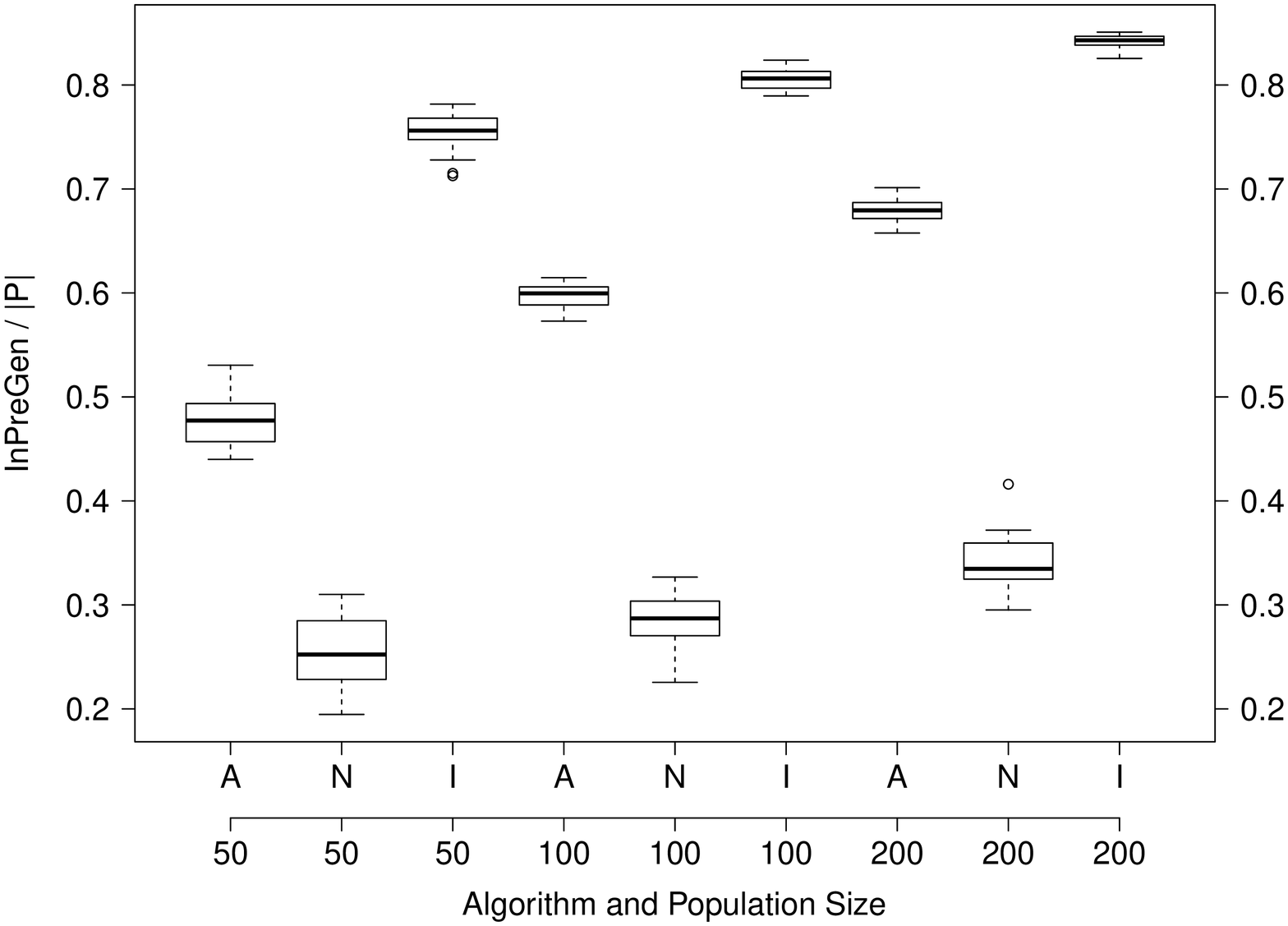}
\begin{center} (b) $\bar {\tau}^-$: Old Pareto optimal \end{center}
\end{minipage}

\begin{minipage}{0.48\textwidth}
  \centering
  \includegraphics[width=0.8\linewidth,keepaspectratio=true, angle=270]{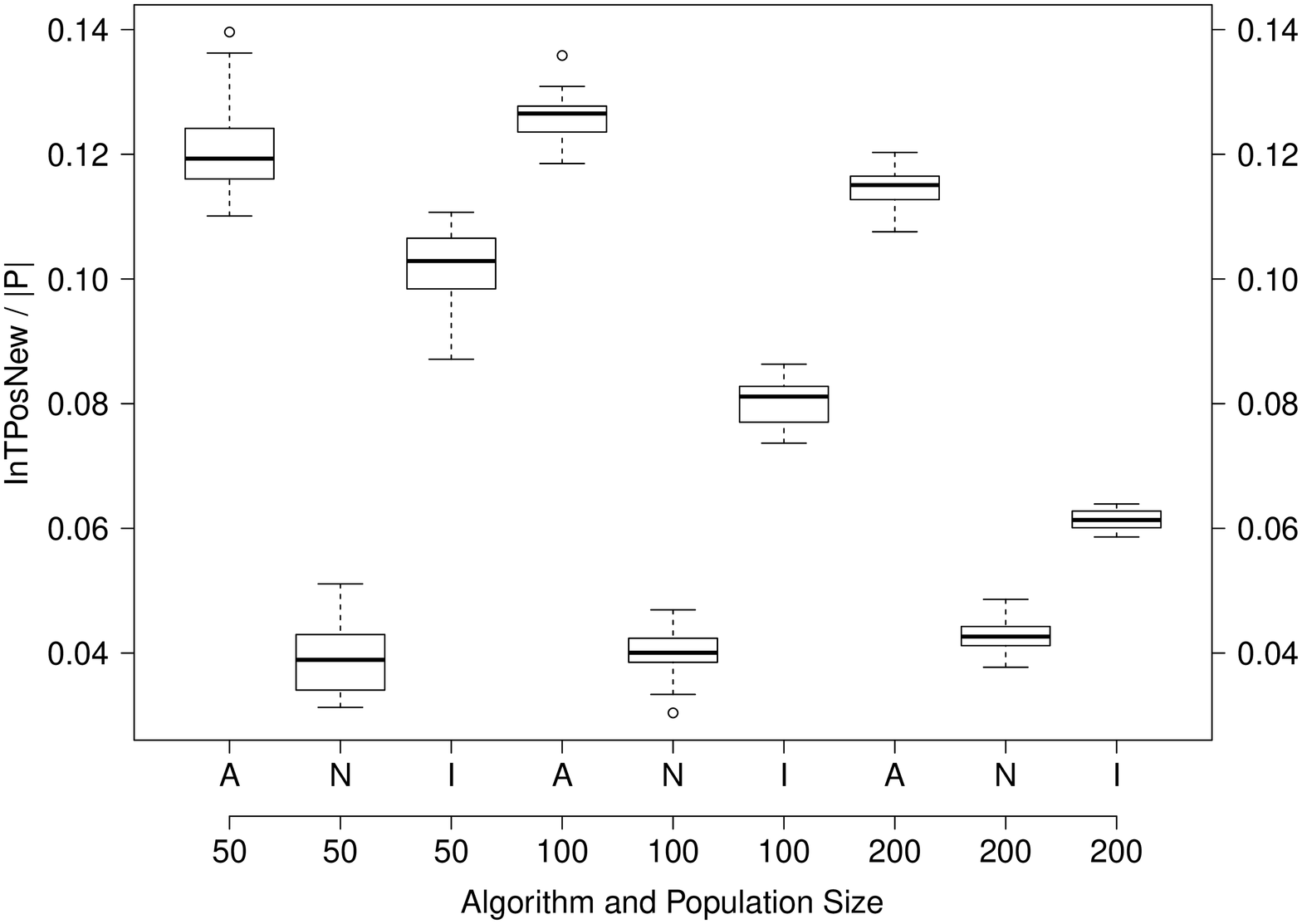}
\begin{center} (c)  $\bar {\tau}^{*}$: Absolutely new Pareto optimal \end{center}
\end{minipage}
\hspace{2mm}
\begin{minipage}{0.48\textwidth}
  \centering
  \includegraphics[width=0.8\linewidth,keepaspectratio=true, angle=270]{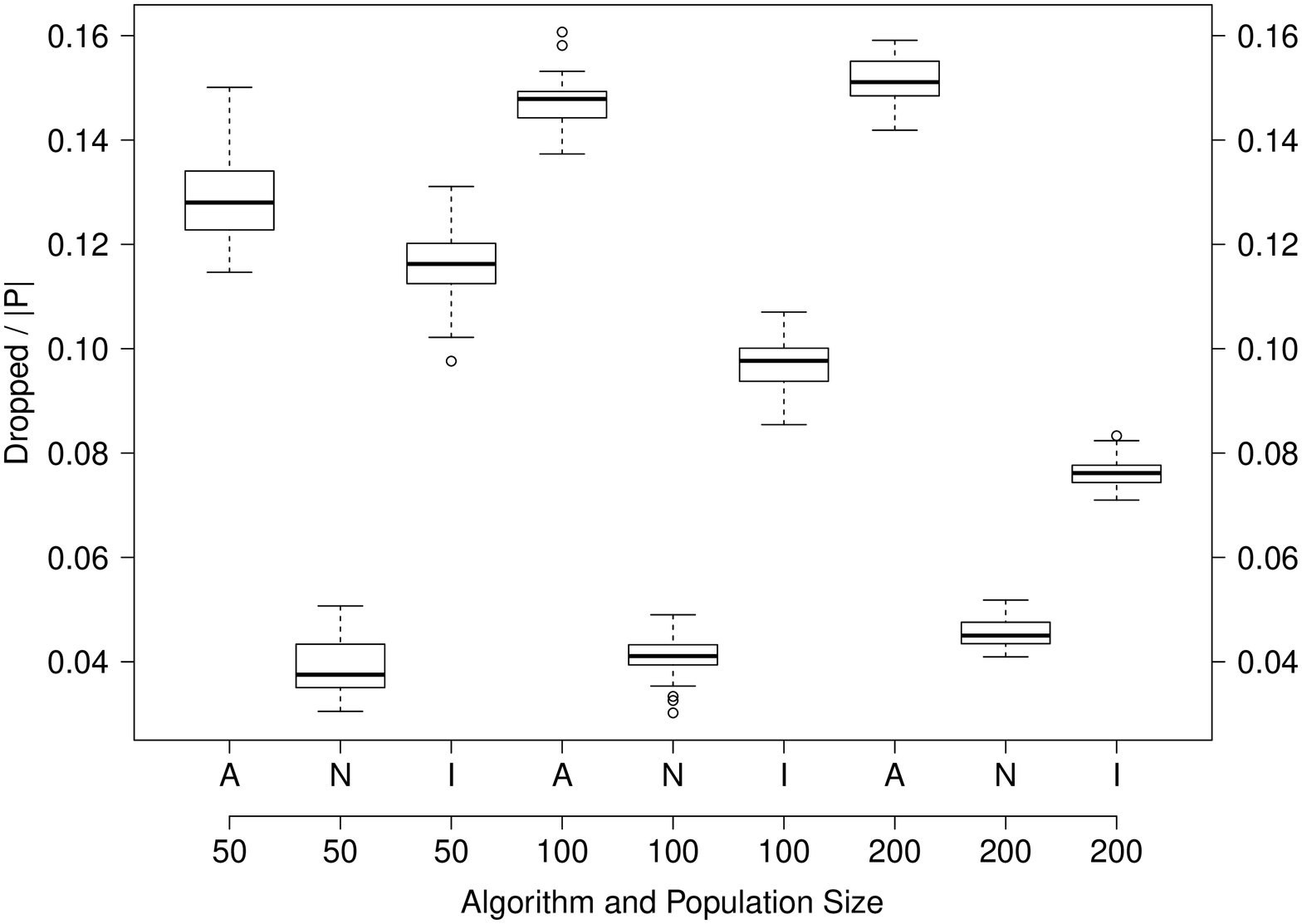}
\begin{center} (d)  $\bar {\delta}$: Dropped Pareto optimal \end{center}
\end{minipage}

  \caption{Boxplots of average generational search-assessment indices in 30 runs. Population sizes $\{50,100, 200\}$,  6 objectives, $T=100$ generations. Algorithms A$\varepsilon$S$\varepsilon$H (A), NSGA-II (N) and IBEA (I). }
  \label{fig:SIp50100200M6G100}
  \vspace{-0.0cm}
\end{figure}

\subsection{Generational search assessment indices}

 \figref{fig:SIp50100200M6G100} (a)-(d) show boxplots of some $\bar{I}$ indexes computed from data obtained in 30 independent runs of the algorithms iterating $T=100$ generations. Results are shown for 6 objectives landscapes using population sizes \{50, 100, 200\}, which are relatively small compared to the POS. From these figures important observations are as follow.
 
In average, at each generation, IBEA contains in its population a very large number of PO solutions compared to  A$\varepsilon$S$\varepsilon$H and NSGA-II as shown in \figref{fig:SIp50100200M6G100} (a). Note that the median of index $\bar{\tau}$ for IBEA is in the range 0.88-0.95, whereas the ranges for  A$\varepsilon$S$\varepsilon$H and NSGA-II are between 0.6-0.85 and 03-0.4, respectively.  

However, the number of old PO solutions (PO solutions present in the current population and also in the population of the previous generation) for IBEA is much larger than for A$\varepsilon$S$\varepsilon$H and NSGA-II, as shown in \figref{fig:SIp50100200M6G100} (b).  Note that the median of $\bar{\tau}^-$ is in the range 0.75-0.85 for IBEA, whereas $\bar{\tau}^-$ is in the range 0.48-0.68 for A$\varepsilon$S$\varepsilon$H and 0.25-0.35 for NSGA-II.

In fact, the generational average number of absolutely new PO solutions (PO solutions in the current population that have not been discovered in previous generations) is larger for A$\varepsilon$S$\varepsilon$H than for IBEA and NSGA-II, as shown in \figref{fig:SIp50100200M6G100} (c). Note that the median of index $\bar{\tau}^*$ for A$\varepsilon$S$\varepsilon$H is around 0.12, whereas for IBEA it reduces with population size from 0.10 to 0.06  and slightly increases for NSGA-II from $0.039$ to  $0.04$. The similar $\bar{\tau}^*$ values by AeSEH are a good sign of robustness to population size variations, i.e. a similar discovery rate could be expected with various population sizes. On the contrary, IBEA's discovery rate could reduce significantly with population size.  If the evaluation of the algorithms is done based only on the points included in the population at a given generation, as it is often the case, IBEA is likely to contain more PO solutions than A$\varepsilon$S$\varepsilon$H, as shown in \figref{fig:SIp50100200M6G100} (a), and therefore be considered a better algorithm.  However,  A$\varepsilon$S$\varepsilon$H finds twice as many PO solutions than IBEA, as shown in \figref{fig:R_AFTPos_TPos_P50100200} (d). These results show that IBEA could be a good algorithm for finding a low resolution approximation of the POS, but for high resolutions is not efficient.  In general, these results show the importance of properly evaluating the algorithms according to the aim of the optimization task at hand. 

The index of dropped PO solutions  $\bar{\delta}$ (PO solutions present in the population of the previous generation that are not included in the current population after truncation selection) shows a trend vey similar to the one observed for the index $\bar{\tau}^*$, as shown in \figref{fig:SIp50100200M6G100} (d).  Dropping superior solutions in favor of solutions that appear non-dominated in the population but are inferior in the landscape could be seen as a selection weakness. However, this could also be a source of exploration. This deserves further research.   

\begin{figure}[t]
\begin{minipage}{0.45\textwidth}
  \centering
  \includegraphics[width=0.8\linewidth,keepaspectratio=true, angle=270]{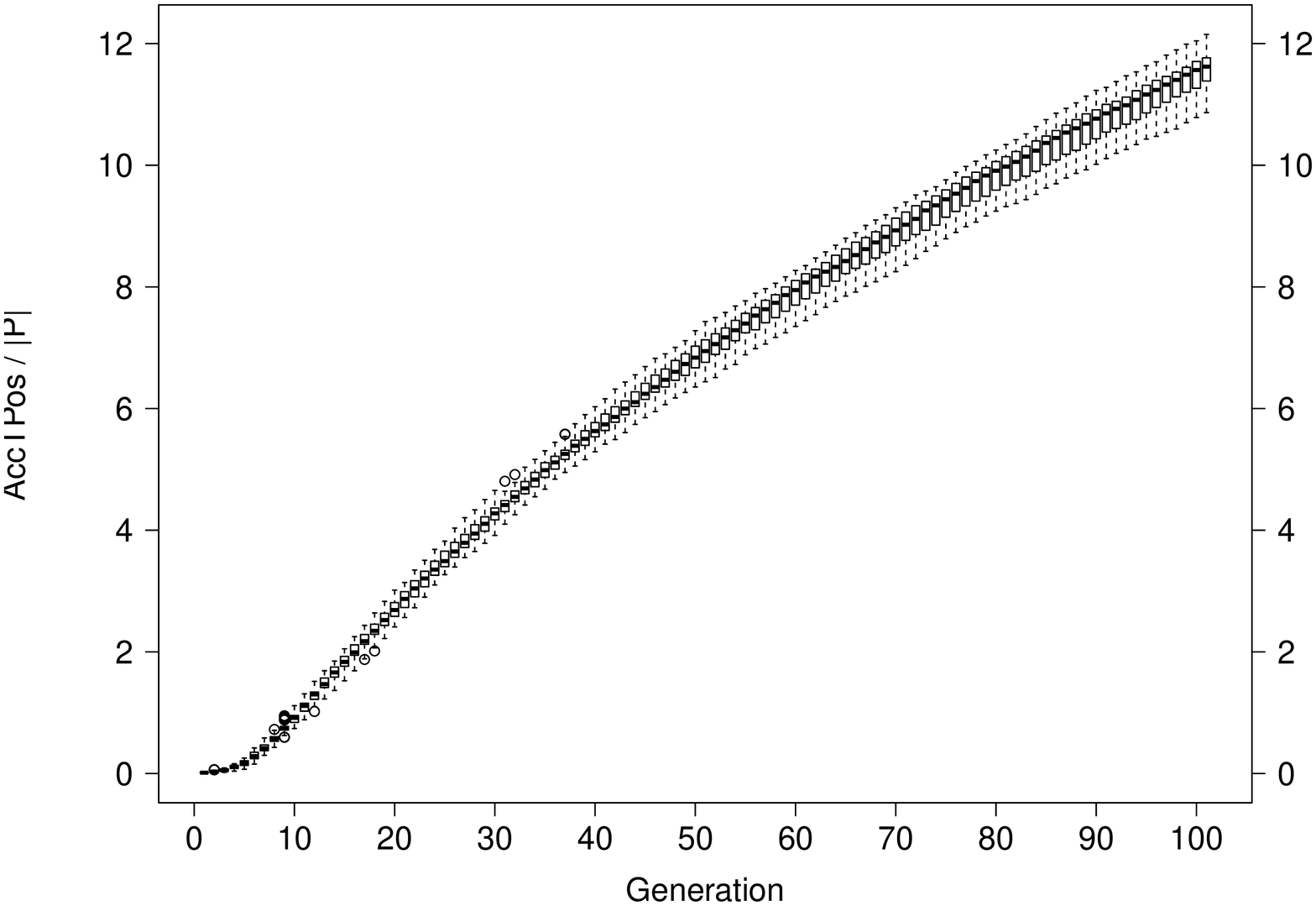}
\begin{center} (a) AeSeH  \end{center}
  \vspace{-0.25cm}
\end{minipage} \hspace{2mm}
\begin{minipage}{0.45\textwidth}
  \centering
  \includegraphics[width=0.8\linewidth,keepaspectratio=true, angle=270]{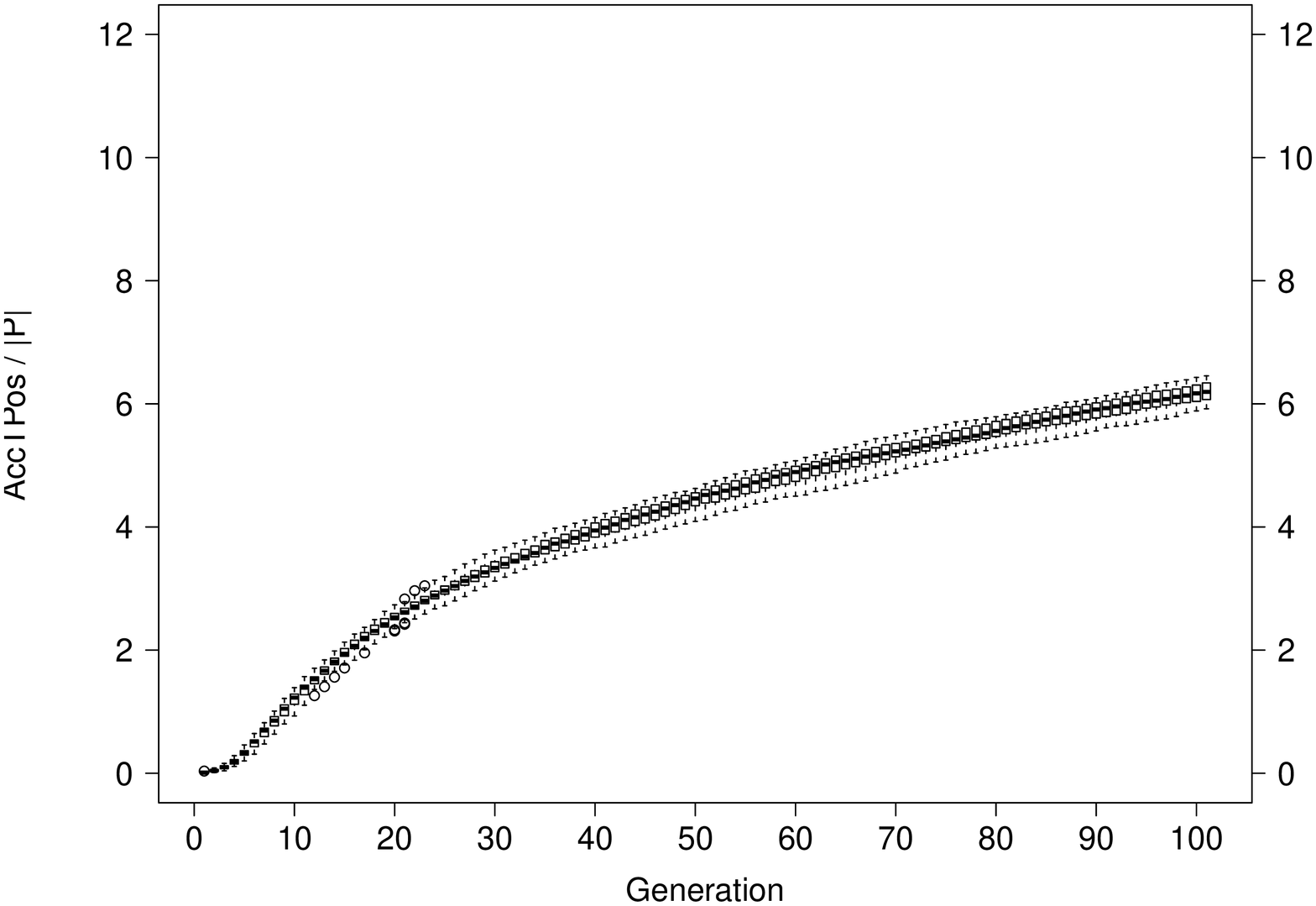}
\begin{center} (b) IBEA  \end{center}
  \vspace{-0.25cm}
\end{minipage} 
  \vspace{-0.25cm}
  \caption{ Accumulated population gains  $\beta(t)$ over the generations. Population size 200, 6 objectives. Algorithms A$\varepsilon$S$\varepsilon$H and IBEA.}
  \label{fig:R_InTPos_P_P200M6}
  \vspace{-0.0cm}
\end{figure}

The accumulated population gains  $\beta(t)$ for  A$\varepsilon$S$\varepsilon$H and IBEA are illustrated in \figref{fig:R_InTPos_P_P200M6} for population size 200 and 6 objectives. Note that just after 20 generation the gain by A$\varepsilon$S$\varepsilon$H is already larger than by IBEA. At the end of the run, A$\varepsilon$S$\varepsilon$H is able to generate an approximation twelve times the size of its population, whereas IBEA  is able to generate an approximation 6 times the size of its population.
  
%------------
\section{Conclusions}

This work has studies the behavior of NSGA-II, IBEA and A$\varepsilon$S$\varepsilon$H generating a high-resolution approximation of the POS. The study has clarified the ability and efficiency of the algorithms assuming scenarios where it is relatively easy to hit the POS, showing the importance to properly assess algorithm's performance according to the task of the optimizer in many objective optimization. In the future, we would like to extend our study to larger landscapes in order to understand the behavior of selection in scenarios where the convergence ability towards the POS is determinant to achieve a good resolution. Also, we would like to study other indicators for IBEA and other many-objective algorithms.

%------------
\balance
%------------------------------------------------------------------------------

\end{document}